\documentclass[conference]{IEEEtran}

\IEEEoverridecommandlockouts
\usepackage{amsmath}
\usepackage{color}
\usepackage{array}
\usepackage{stfloats}
\usepackage{amsthm} 
\usepackage{amssymb}
\usepackage{float}
\usepackage{nccmath}
\usepackage{graphicx}
\usepackage[linesnumbered,ruled]{algorithm2e}
\usepackage{setspace}
\usepackage{booktabs}
\usepackage{subcaption}
\usepackage{mwe}
\usepackage{cite}
\usepackage{amsmath,amssymb,amsfonts}
\usepackage{graphicx}
\usepackage{textcomp}
\usepackage{xcolor}
\usepackage[utf8]{inputenc} 
\usepackage{kotex} 

\usepackage{algorithmicx}
\usepackage{xspace}
\def\BibTeX{{\rm B\kern-.05em{\sc i\kern-.025em b}\kern-.08em
    T\kern-.1667em\lower.7ex\hbox{E}\kern-.125emX}}
\begin{document}
\newcommand{\BfPara}[1]{{\noindent\bf#1.}\xspace}
\newcommand\mycaption[2]{\caption{#1\newline\small#2}}
\newcommand\mycap[3]{\caption{#1\newline\small#2\newline\small#3}}

\title{Spatio-Temporal Split Learning}

\author{
\IEEEauthorblockN{Joongheon Kim}
\IEEEauthorblockA{\textit{Korea University}\\
Seoul, Korea \\
joongheon@korea.ac.kr}
\and
\IEEEauthorblockN{Seunghoon Park}
\IEEEauthorblockA{\textit{Korea University} \\
Seoul, Korea \\
psh95king@korea.ac.kr}
\and
\IEEEauthorblockN{Soyi Jung}
\IEEEauthorblockA{\textit{Korea University} \\
Seoul, Korea \\
jungsoyi@korea.ac.kr}
\and
\IEEEauthorblockN{Seehwan Yoo}
\IEEEauthorblockA{\textit{Dankook University}\\
Yongin, Korea \\
seehwan.yoo@dankook.ac.kr}
}

\maketitle

\begin{abstract}
This paper proposes a novel split learning framework with multiple end-systems in order to realize privacy-preserving deep neural network computation. 
In conventional split learning frameworks, deep neural network computation is separated into multiple computing systems for hiding entire network architectures. 
In our proposed framework, multiple computing end-systems are sharing one centralized server in split learning computation, where the multiple end-systems are with input and first hidden layers and the centralized server is with the other hidden layers and output layer. This framework, which is called as \textit{spatio-temporal split learning}, is spatially separated for gathering data from multiple end-systems and also temporally separated due to the nature of split learning. Our performance evaluation verifies that our proposed framework shows near-optimal accuracy while preserving data privacy.
\end{abstract}


\section{Introduction}
\begin{figure}[t!]
\centerline{\includegraphics[width=85mm]{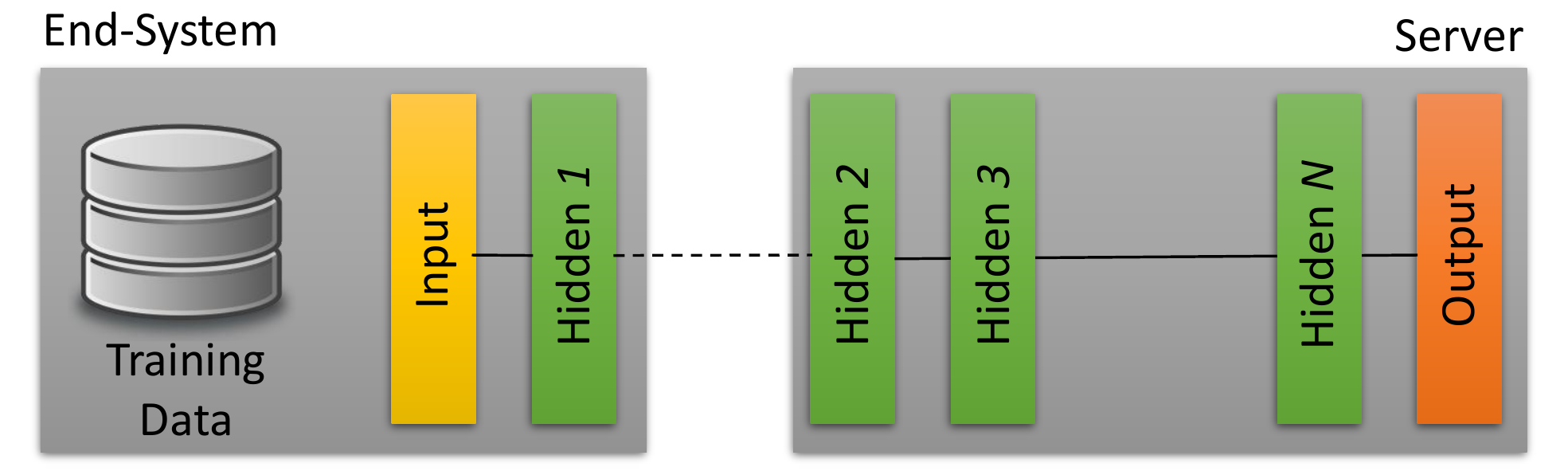}}
\caption{Basic system architecture for split learning.}
\vspace{-3mm}
\label{fig:fig1}
\end{figure}

Nowadays, deep neural network computations are widely and actively used in many applications. 
Therefore, it is obvious that there are a lot of research contributions which aim at privacy-preserving secure deep neural network computation. 
Among them, federated learning is one of well-known and successful approaches which does not expose training data to the public~\cite{pieee21park,iotj20kwon}. 
According to the privacy-preserving nature in federated learning, it is suitable for medical applications. 
In distributed medical systems (hospitals,  biomedical research institutes, etc.), there exist lots of patients' data. In addition, the system conducts deep neural network training on such data all of it typically should be in the same space for the training. However, gathering and transporting patient data is strictly regulated by laws due to privacy. As a result, each medical system conducts computations with its own local data. To deal with this issue, this paper proposes a federated learning framework which trains models with the data stored in each end-system \textit{locally} for privacy-preserving computation while maintaining learning accuracy. 
Among various federated learning algorithms, this paper considers split learning which separates deep neural network computation into several parts~\cite{splitlearning}. 
The basic system architecture for split learning is as illustrated in Fig.~\ref{fig:fig1}~\cite{dsn19jeon}.
As shown in Fig.~\ref{fig:fig1}, the training data located in end-system cannot be seen at the centralized server, thus, privacy-preserving  deep neural network computation can be available. 
However, to the best of our knowledge, multiple end-systems are not considered in split learning research contributions, yet. In the case with multiple end-systems, the training data are located in spatially separated individual end-systems and the deep neural network training is temporally separated as shown in Fig.~\ref{fig:fig1}, thus our proposed system is named to \textit{spatio-temporal split learning}. 
In this proposed spatio-temporal split learning, each end-system contains several hidden layers and the results of the last hidden layers of all end-systems will be delivered to the centralized server. The server has the other remaining hidden layers and output layer of our considering deep neural network, thus, centralized computation can be available.
Based on our performance evaluation results with cifar10 based classification, it can be observed that our spatio-temporal split learning can show near-optimal performance while preserving data privacy. 

\section{Spatio-Temporal Split Learning Framework}

\begin{figure}[t!]
\centerline{\includegraphics[width=85mm]{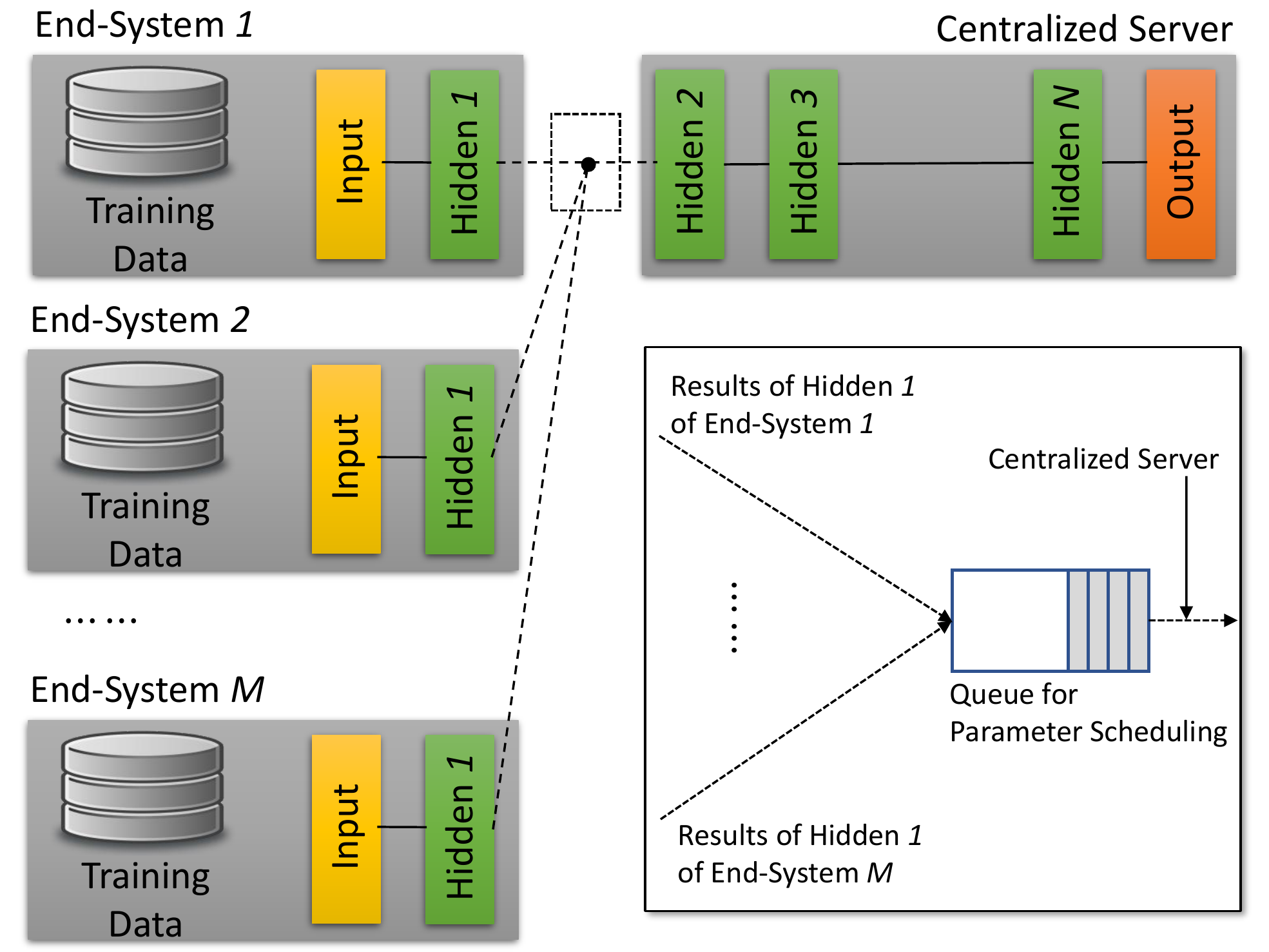}}
\caption{Our proposed spatio-temporal split learning framework.}
\label{fig:fig2}
\end{figure}

Our proposed spatio-temporal split learning framework is as illustrated in Fig.~\ref{fig:fig2}. As shown in Fig.~\ref{fig:fig2}, multiple end-systems exist; and the individual results of end-systems' first hidden layers are delivered to the centralized server. Then, the original raw data is not shared and encoded (due to the first hidden layer computation). Note that sharing the results of first hidden layers at the centralized server does not expose original raw data. Then, deep neural network computation with the other hidden layers and output layer is conducted at the server, thus, all training data is used for single deep neural network training. Therefore, in theory, our proposed spatio-temporal split learning framework achieves near optimal performance by using all data in a single network with certain amount of performance sacrifice 
due to individual first hidden layers in each end-system. This performance degradation can be larger when more hidden layers are in end-systems while sacrificing certain amounts of learning performance (tradeoff). 

The centralized server requires queue while gathering the results of the fist hidden layers in end-systems under the consideration of geo-distributed end-systems. If an end-system is located very far from the centralized server, the parameters from the end-system can arrive at the server lately or sparsely. Then, the learning performance can be biased due to the differences of arrivals from end-systems. Thus, parameter scheduling is required depending on applications, i.e., a queue data structure needs to be defined.

\begin{figure}[t!]
\centerline{\includegraphics[width=85mm]{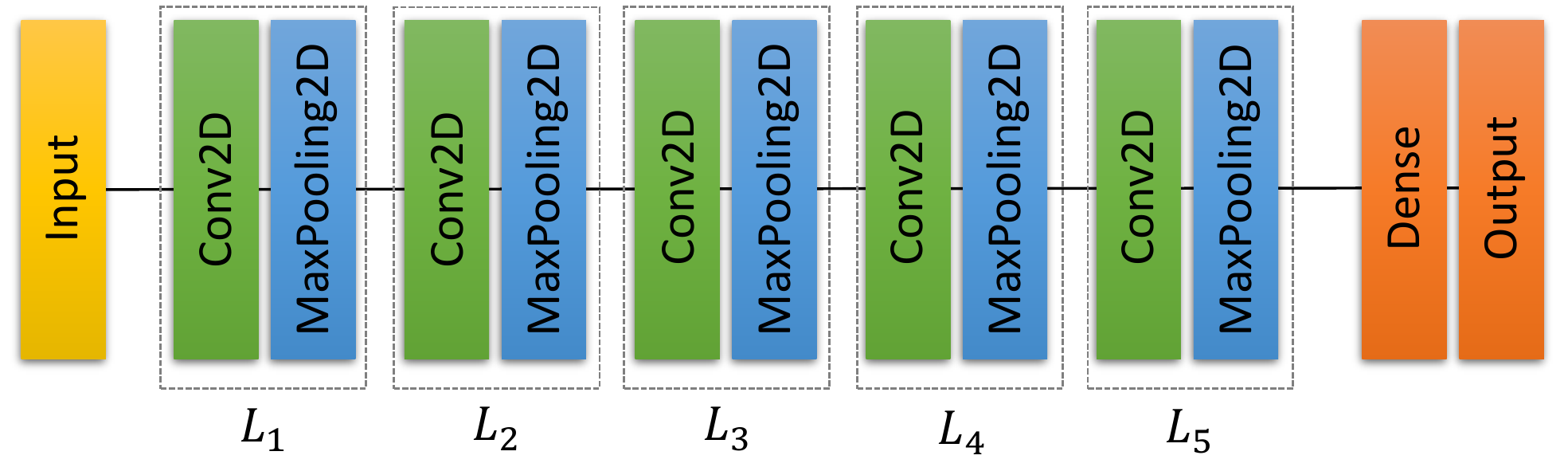}}
\caption{Our considering CNN for cifar10 classification CNN for performance evaluation.}
\label{fig:fig3}
\end{figure}

\begin{table}[t!]%
\caption{Accuracy Results}
\label{accuracy}
\small
\begin{center}
	\begin{tabular}{r|r}
    \toprule[1.0pt]
    Layers at end-systems & Accuracy \\
    \midrule
    Nothing (All layers are in the server) & $71.09$\,\% \\
    $L_{1}$ & $68.18$\,\% \\
    $L_{1}, L_{2}$ & $67.92$\,\% \\
    $L_{1}, L_{2}, L_{3}$ & $66.00$\,\% \\
    $L_{1}, L_{2}, L_{3}, L_{4}$ & $65.66$\,\% \\
    \bottomrule[1.0pt]
	\end{tabular}
\end{center}
\vspace{-5mm}
\end{table}

\section{Primary Evaluation Results}
In order to show that our proposed spatio-temporal split learning framework presents near-optimal performance while preserving data privacy, cifar10 based classification is conducted with convolutional neural networks (CNN). 
In our considering CNN, as illustrated in Fig.~\ref{fig:fig3}, 5 convolution layers (Conv2D) and 5 max-pooling layers (MaxPooling2D) are used.
In $L_{1},\cdots,L_{5}$ (Conv2D and MaxPooling2D) layers, $16$, $32$, $64$, $128$, and $256$ numbers of filters are used where the size is $32\times 32$. The last two dense layers (including output layer) has $512$ and $10$ units. 
Then, our performance evaluation is conducted while the numbers of $L_{i}, \forall i\in\{1,\cdots,5\}$ in end-systems vary, i.e., from $L_{1}$ to $L_{5}$.  

\begin{figure}[t!]
\centering
\setlength{\tabcolsep}{2pt}
\renewcommand{\arraystretch}{0.2}
\begin{tabular}{p{0.31\linewidth}p{0.02\linewidth}p{0.31\linewidth}p{0.02\linewidth}p{0.31\linewidth}}
\tabularnewline
\tabularnewline
\includegraphics[page=1, width=0.95\linewidth]{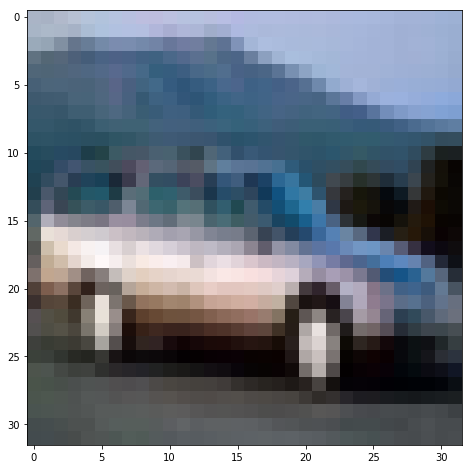} & {} &
\includegraphics[page=1, width=0.95\linewidth]{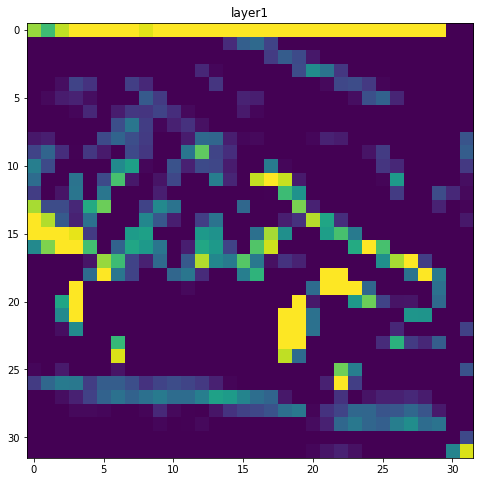} & {} &
\includegraphics[page=1, width=0.95\linewidth]{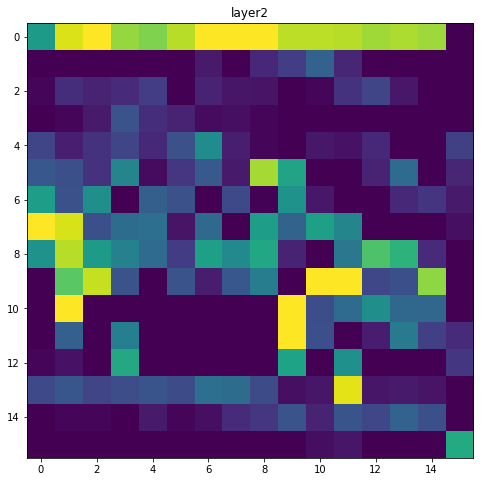}
\\
\tabularnewline
\tabularnewline
\centering(a) Original & {} &
\centering(b) Conv2D in $L_{1}$ & {} &
\centering(c) $L_{1}$
\tabularnewline
\end{tabular}
\caption{Image capture during deep neural network computation.} 
\label{fig:privacy}
\vspace{-5mm}
\end{figure}

As shown in Table~\ref{accuracy}, $71.09$\,\% classification accuracy can be obtained if all layers are in the centralized server (global model). If $L_{1}$ is located in end-systems, the performance is degraded from $71.09$\,\% to $68.18$\,\%, i.e., $2.91$\,\%. With this small amount of performance degradation, original raw data at end-systems are not exposed/shared, thus, the privacy of training data is preserved. In the worst case in our experiments where $L_{1},\cdots,L_{4}$ layers are in end-systems, the performance is $65.66$\,\%, thus, we only have $5.43$\,\% performance degradation. It means that our proposed spatio-temporal split learning works well.
As shown in Fig.~\ref{fig:privacy}, original training cifar10 images may be recognized only with the Conv2D in $L_{1}$ (even if they are blurred) as in Fig.~\ref{fig:privacy}(b), however, max-pooling can definitely hide original images, as in Fig.~\ref{fig:privacy}(c).

\section{Summary and Future Work}
In this paper, a novel privacy-preserving split learning framework with multiple end-systems, called as \textit{spatio-temporal split learning}, is proposed for avoiding original raw data sharing among end-systems during deep neural network training. 
In our proposed framework, multiple end-systems are sharing one centralized server, where the multiple end-systems are with first hidden layer and the centralized server is with the other layers. This framework is spatially separated for getting data from multiple end-systems and temporally separated for split learning computation. The performance of the proposed framework is evaluated and the results verify that 
the framework shows near-optimal performance without original raw data sharing for privacy preserving computation.

\section*{Acknowledgment}
This research was funded by Ministry of Health and Welfare (HI19C0572) and National Research Foundation of Korea (2019R1A2C4070663). 
S. Jung and S. Yoo are corresponding authors.

\begin{thebibliography}{00}

\bibitem{pieee21park}
J. Park, S. Samarakoon, A. Elgabli, J. Kim, M. Bennis, S. Kim, and M. Debbah, ``Communication-efficient and distributed learning over wireless networks: Principles and applications," \textit{Proc. of the IEEE}, 2021.

\bibitem{iotj20kwon}
D. Kwon, J. Jeon, S. Park, J. Kim, and S. Cho, ``Multiagent DDPG-based deep learning for smart ocean federated learning IoT networks," \textit{IEEE Internet of Things J.}, vol. 7, no. 10, pp. 9895–9903, Oct. 2020.

\bibitem{splitlearning}
P. Vepakomma, O. Gupta, T. Swedish, and R. Raskar, ``Split learning
for health: Distributed deep learning without sharing raw patient data," \textit{CoRR}, vol. abs/1812.00564, 2018.

\bibitem{dsn19jeon}
J. Jeon, J. Kim, J. Kim, K. Kim, A. Mohaisen, and J. Kim, ``Privacy-preserving deep learning computation for geo-distributed medical big-data platforms," in \textit{Proc. IEEE/IFIP International Conference on Dependable Systems and Networks (DSN)}, Portland, OR, USA, June 2019.

\end{thebibliography}
\end{document}